\documentclass[letterpaper, 10 pt, journal, twoside]{IEEEtran}

\IEEEoverridecommandlockouts                              

\usepackage{amsmath,amsfonts}
\usepackage{bm}
\usepackage[table,xcdraw]{xcolor}
\usepackage{url}
\usepackage{graphicx}
\usepackage{mathtools}
\usepackage{dirtree}
\usepackage{amssymb} 
\usepackage{breqn}
\usepackage{multirow}
\usepackage{cancel} 
\usepackage{xfrac}
\usepackage{xspace}
\usepackage{booktabs}
\usepackage{pifont}
\usepackage{arydshln}
\usepackage{makecell}  
\usepackage{balance}
\usepackage{verbatim}
\usepackage{comment}
\makeatletter
\let\NAT@parse\undefined
\makeatother
\usepackage[colorlinks=true,
			linkcolor=blue,
			urlcolor=blue,
			citecolor=blue]{hyperref}

\newcommand{\parenthesis}[1]{
	\left(#1\right)
}

\DeclareMathOperator*{\argmin}{arg\hspace{0.5mm}min\hspace{1mm}}

{\endarray\hspace*{-0.5\arraycolsep}\endBmatrix}
\newenvironment{bMatrix}[1]{%
\bmatrix\array{#1}\hspace*{-0.5\arraycolsep}}%
{\endarray\endbmatrix}

\newcommand{\T}{\mathbf{T}}

\usepackage{balance}

\begin{document}

\title{GNSS-Inertial State Initialization Using Inter-Epoch Baseline Residuals}

\author{Samuel Cerezo$^1$, Javier Civera$^1$
\thanks{Manuscript received: June, 12, 2025; Revised October, 3, 2025; Accepted November, 30, 2025.}
\thanks{This paper was recommended for publication by Editor Pascal Vasseur upon evaluation of the Associate Editor and Reviewers' comments.
This work was supported by the Spanish Government (PID2021-127685NB-I00, TED2021-131150B-I00) and the Agencia Estatal de Investigación / Ministerio de Ciencia e Innovación (AEI/MCIN) under grant PRE2022-103765.} 
\thanks{$^1$The authors are with School of Engineering and Architecture,
        Universidad de Zaragoza, C/María de Luna, 1. 50018, Zaragoza
        {\tt\small \{sacerezo,jcivera\}@unizar.es}}
\thanks{Digital Object Identifier (DOI): see top of this page.}
}

\markboth{IEEE Robotics and Automation Letters. Preprint Version. Accepted November, 2025}
{Cerezo \MakeLowercase{\textit{et al.}}: GNSS-Inertial State Initialization Using Inter-Epoch Baseline Residuals}

\maketitle


\begin{abstract}

Initializing the state of a sensorized platform can be challenging, as a limited set of measurements often provide low-informative constraints that are in addition highly non-linear. This may lead to poor initial estimates that may converge to local minima during subsequent non-linear optimization. 
We propose an adaptive GNSS–inertial initialization strategy that delays the incorporation of global GNSS constraints until they become sufficiently informative. 
In the initial stage, our method leverages inter-epoch baseline vector residuals between consecutive GNSS fixes to mitigate inertial drift. 
To determine when to activate global constraints, we introduce a general criterion based on the evolution of the Hessian matrix’s singular values, effectively quantifying system observability. 
Experiments on EuRoC, GVINS and MARS-LVIG  datasets show that our approach consistently outperforms the naive strategy of fusing all measurements from the outset, yielding more accurate and robust initializations.
\end{abstract}

\begin{IEEEkeywords}
SLAM, Sensor Fusion
\end{IEEEkeywords}

\section{Introduction}
\label{sec:introduction}
\IEEEPARstart{S}{tate} initialization is a well-known challenge in sensor fusion, as evidenced by extensive research in the visual-inertial domain~\cite{martinelli2014closed,campos2020inertial,he2023rotation}. Early sensor measurements often carry limited information, and certain states may even be unobservable. As a result, initial estimates can be highly inaccurate, leading to convergence toward local minima rather than the global optimum. Although this problem has been widely studied in the visual-inertial domain, analogous challenges may arise in other sensor fusion modalities.

In this paper, we demonstrate that state initialization is equally critical in GNSS–inertial systems, where naive strategies may cause inconsistent estimates. We propose a two-stage initialization method that postpones the use of global position measurements until they become sufficiently informative. To determine the appropriate timing, we derive a general criterion that evaluates the informativeness of the available data for accurate state estimation. During the initial stage, the method relies solely on inter-epoch baseline vectors
between GNSS measurements, effectively mitigating inertial preintegration drift. Experimental results show that our approach consistently outperforms the naive strategy of fusing all measurements from the outset.
Note that we do not target steady-state navigation~\cite{Syed2008CivilianVehicle}, but producing robust and accurate initialization estimates to reliably bootstrap UAVs and other mobile robots in field conditions.

The complementary nature of GNSS and inertial sensors motivates the need for such an initialization strategy. GNSS provides global positioning capabilities fundamental to modern navigation, but its accuracy can degrade under multipath, atmospheric effects, or obstructed satellite visibility~\cite{tang2018gnss}. Inertial sensors, by contrast, offer high-frequency motion cues that are immune to external disturbances, but their integration inevitably leads to cumulative drift. Moreover, under certain motion patterns—such as constant velocity or straight-line trajectories—inertial states may become only partially observable. These limitations underscore the importance of principled initialization methods that can exploit early measurements effectively and ensure robust fusion performance.

\begin{figure}[]
    \centering
    {\includegraphics[width=0.93\linewidth]{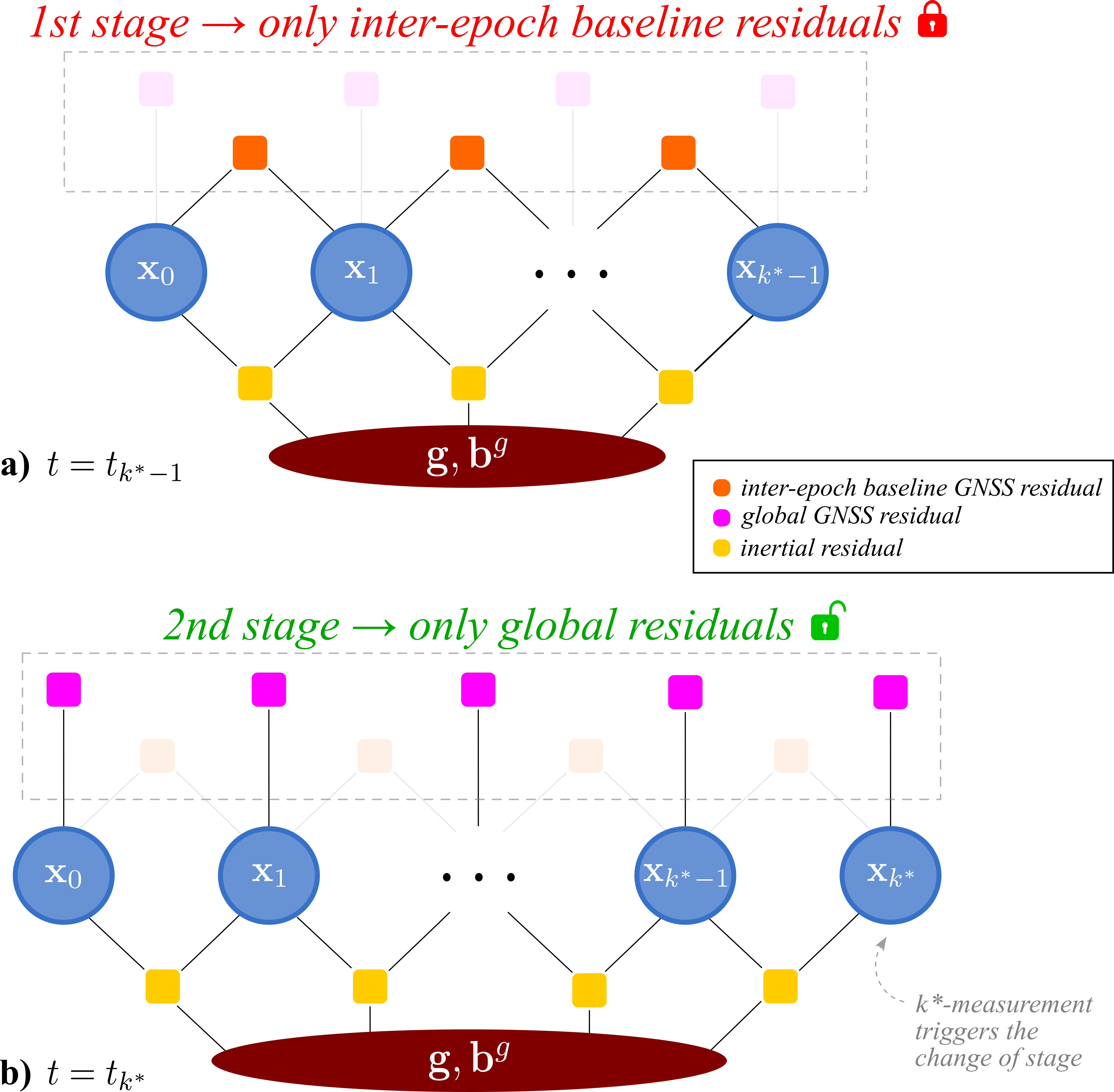}}
   \caption{Illustration of our GNSS-inertial state initialization. a) Global residuals remain inactive while the system lacks informative measurements. b) At time $k^*$, the GNSS-inertial measurements provide sufficient information to meet our criterion, enabling the inclusion of global constraints for accurate alignment.
    }
  \label{fig:grafo-gpsi}
\end{figure}

\section{Related Work}
\label{sec:related-work}


\subsection{GNSS-inertial alignment}

Several methods have been proposed for GNSS-inertial alignment, particularly in scenarios involving low-cost MEMS sensors and challenging motion conditions. Shin and El-Sheimy~\cite{shin2004ukf} introduced an Unscented Kalman filter (UKF) framework for in-motion alignment, capable of handling large initial attitude uncertainties without stationary initialization. Zhang et al.~\cite{zhang2020vboba} later proposed a velocity-based optimization-based alignment (VBOBA) method that exploits GNSS velocity vectors to improve heading observability under low-dynamic conditions. Groves~\cite{groves2008} provides a comprehensive treatment of transfer alignment techniques, where a high-grade reference system supports the initialization of a secondary INS in motion. Wu and Pan~\cite{Wu_VelPosIntegration} further derived the velocity/position integration formula, laying theoretical foundations for subsequent optimization-based alignment techniques. 
These approaches remain highly relevant and complementary. In contrast, our method is designed to guarantee robust initialization that prevents divergence, rather than aiming at fine-grained in-motion alignment, and thus provides a different but compatible contribution to the GNSS-inertial initialization literature.
Wu et al.~\cite{wu2014new} introduced RA-JAPE, a time-recursive Newton–Lagrange optimization that jointly estimates attitude, IMU biases, and the GNSS lever arm, enabling self-initialization and outperforming state-of-the-art methods.

\subsection{GNSS--IMU localization beyond filtering}
In the absence of vision or contextual information, GNSS-inertial fusion remains crucial, particularly in cost-sensitive or resource-constrained applications.
He et al.~\cite{he2014gnss} developed an adaptive robust Kalman filter that suppresses outliers in GNSS signals, ensuring stable integration in scenarios with signal degradation or interference. Kim et al.~\cite{kim2017compressed} addressed computational efficiency by a compressed fusion strategy that embeds GNSS-inertial fusion within a SLAM-like architecture, maintaining reliable estimates with reduced complexity.
Zhang et al.~\cite{zhang2019kgp} proposed the KGP algorithm, which enhances error-state updates in INS/GNSS integration through improved modeling, resulting in a more accurate and stable positioning across dynamic navigation tasks.

Beyond linear error-state filtering, factor-graph optimization (FGO) offers a flexible backend to fuse IMU preintegration with GNSS observations (code, carrier, Doppler), often yielding higher accuracy and resilience under degraded satellite geometry~\cite{forster-manifold-2017,dellaert2017factor}. Robustness to outliers and mis-modeling can be handled with robust costs and \emph{switchable constraints}, which softly activate/deactivate factors based on consistency---a mechanism closely related to observability-aware gating~\cite{sunderhauf2012switchable}.

Unlike methods based on linear models and filtering, our approach formulates the problem as a nonlinear optimization, providing increased flexibility and accuracy in modeling system dynamics. In the early stages, the formulation includes 
inter-epoch baseline vectors (i.e., consecutive GNSS position differences)
which ensure geometric consistency without requiring global alignment. Once full observability is achieved, these relative residuals are deactivated and replaced by absolute position residuals, enabling direct alignment to the global frame. This switch is governed by an observability-based mechanism that ensures global constraints are introduced only when the estimation problem is sufficiently well-conditioned, thereby improving trajectory consistency and overall robustness.

\section{Notation and background}
\label{sec:notation-&-background}

Throughout this article, bold lower-case letters ($\mathbf{x}$) represent vectors and bold upper-case letters ($\boldsymbol{\Sigma}$) represent matrices. Scalars will be denoted as light lower-case letters (${\alpha}$) and functions by light upper-case letters ($J$).
Six-degrees-of-freedom motion will be represented by the rotation matrix $\mathbf{R}_{WB} \in SO(3)$ and translation vector $^{W}\mathbf{p} \in \mathbb{R}^3$.

Optimization is carried out on the tangent spaces $\mathfrak{so}(3)$ and $\mathfrak{se}(3)$ of $SO(3)$ and $SE(3)$, respectively, using the exponential and logarithmic maps for retraction-based updates~\cite{sola_micro_2018}.
We use the hat operator $(\cdot)^\wedge:\mathbb{R}^3\!\to\!\mathfrak{so}(3)$, which maps a 3D vector to its skew-symmetric matrix, and the vee operator $(\cdot)^\vee:\mathfrak{so}(3)\!\to\!\mathbb{R}^3$, its inverse.
Closed-form expressions for the exponential/logarithmic maps and the inverse right Jacobian of $SO(3)$, required for linearizing rotational residuals, are detailed in~\cite{forster-manifold-2017}.

\subsection{IMU dynamics}
\label{subsec:IMU-model}
An IMU consists typically of an accelerometer and a three-axis gyroscope, and measures the angular velocity $\boldsymbol{\omega}$ and linear acceleration of the sensor $\mathbf{a}$ with respect to a local inertial frame.
We will denote the IMU measurement at time instant $k$ as $^{B}\Tilde{\boldsymbol{\omega}}_{k}$ and $^{B}\Tilde{\mathbf{a}}_k$. These measurements are affected by additive noise $\boldsymbol{\eta}^{g}$, $\boldsymbol{\eta}^{a} \in \mathbb{R}^{3}$ and two slowly varying sensor biases $\mathbf{b}^g$ and $\mathbf{b}^a \in \mathbb{R}^{3}$, which are the gyroscope and accelerometer bias, respectively. These model is formulated by (\ref{eq:imu_model1}) and (\ref{eq:imu_model2}). 
The prefix $^{B}\cdot$ denotes that the corresponding magnitude is expressed in the body reference frame $B$.
\begin{equation}
\label{eq:imu_model1}
        ^{B}\Tilde{\boldsymbol{\omega}}_k =\hspace{0.5mm} ^{B}\boldsymbol{\omega}_k + \mathbf{b}^g_k + \boldsymbol{\eta}^{g}_k
\end{equation}
\begin{equation}
\label{eq:imu_model2}
        ^{B}\Tilde{\mathbf{a}}_k = \mathbf{R}_{WB}^{\top}\left(^{W}\mathbf{a}_k -\hspace{0.5mm} ^{W}\mathbf{g}\right) + \mathbf{b}^a_k + \boldsymbol{\eta}^{a}_k
\end{equation}

\noindent Using relative motion increment $ \Delta \mathbf{v}_{ij} \doteq \hspace{0.5mm} \mathbf{R}_i^{\top}(\mathbf{v}_j - \mathbf{v}_i - \mathbf{g}\Delta t_{ij}),$ a gravity estimation can be obtained
\begin{equation}
{\mathbf{g}}^* = \frac{\mathbf{v}_j - \mathbf{v}_i}{\Delta t_{ij}} - \frac{\mathbf{R}_i\Delta \mathbf{v}_{ij}}{\Delta t_{ij}}.  \end{equation}

\subsection{Noise propagation}
The covariance matrix of the raw IMU measurements noise $\boldsymbol{{\Sigma}_{\eta}} \in \mathbb{S}^{6}_+$ is composed\footnote{By $\mathbb{S}^{n}_+=\{\mathbf{\Sigma} \in \mathbb{R}^{n \times n} \ | \ \mathbf{\Sigma} = \mathbf{\Sigma}^\top, \mathbf{\Sigma}\succeq 0\}$ we denote the set of $n \times n$ symmetric positive semidefinite matrices.} by sub-matrices
$\boldsymbol{{\Sigma}_{\omega}},\boldsymbol{{\Sigma}_{\textbf{a}}} \in \mathbb{S}^{3}_+$
\begin{equation}
\label{eq:matriz-covarianza-ruido}
 \boldsymbol{{\Sigma}_{\eta}}
  =\begin{bMatrix}{c:c}
  \boldsymbol{{\Sigma}_{\omega}} & \textbf{0}_{3\mathrm{x}3}\\ \hdashline \textbf{0}_{3\mathrm{x}3} & \boldsymbol{{\Sigma}_{\textbf{a}}}
  \end{bMatrix}
\end{equation}
\noindent where $\textbf{0}_{3\mathrm{x}3}$ is a $3\mathrm{x}3$ matrix for which all its elements are equal to zero.

Following the computation of the preintegrated noise covariance in \cite{forster-manifold-2017}, we consider the matrix $\boldsymbol\eta_{ik}^{\Delta} \doteq \left[\delta \boldsymbol\phi_{ik}^\top , \delta \mathbf{v}_{ik}^\top , \delta   \mathbf{p}_{ik}^\top\right]^\top$
and defining the IMU measurement noise $\boldsymbol\eta_{k}^{d} \doteq \left[\boldsymbol\eta_{k}^{gd} , \boldsymbol\eta_{k}^{ad}\right]^\top$, 
the noise is propagated as
\begin{equation}
\label{eq:def-preintegrated-measurement-covariance}
        {\boldsymbol\Sigma}_{ij} = \mathbf{A}_{j-1}{\boldsymbol\Sigma}_{ij-1}\mathbf{A}_{j-1}^\top +  \mathbf{B}_{j-1}\boldsymbol{{\Sigma}_{\eta}}\mathbf{B}_{j-1}^\top,
\end{equation}
\noindent with initial conditions ${\boldsymbol\Sigma}_{ii} = \textbf{0}_{9\times 9}$.
Matrices $\mathbf{A}_{j-1} \in \mathbb{R}^{9\times 9}$ and $\mathbf{B}_{j-1} \in \mathbb{R}^{9\times6}$, together with the corresponding covariance propagation rules, are expressed explicitly in~\cite{camera-motion2024}. 
The noise matrix ${\boldsymbol\Sigma}_{ij} \in \mathbb{S}^{9}_+$ captures the uncertainty of rotation, velocity, and position increments and is a key component in defining the residuals in the next section.

\subsection{Gravity direction representation}
\label{subsec:gravity-representation}

Since the gravity magnitude is known, an appropriate representation is through its direction vector. Unit-norm vectors lie on the sphere manifold $\mathbb{S}^{2}$, which has only two degrees of freedom. As $\mathbb{S}^{2}$ does not form a Lie group, we adopt the local perturbation model proposed in~\cite{hertzberg2013integrating}. Given unit vectors $\hat{\mathbf{g}}_i \in \mathbb{S}^{2}$ and a local perturbation $\boldsymbol{\delta} \in \mathbb{R}^{2}$, the $\boxplus$ and $\boxminus$ operators are defined as follows
\begin{equation*}
    \hat{\mathbf{g}}_1 \boxplus \boldsymbol{\delta} = \mathbf{R}_{\hat{\mathbf{g}}_1} \, \mathrm{Exp}_{\mathbb{S}^2}(\boldsymbol{\delta})
    ,\quad 
    \hat{\mathbf{g}}_2 \boxminus \hat{\mathbf{g}}_1 = \mathrm{Log}_{\mathbb{S}^2}(\mathbf{R}_{\hat{\mathbf{g}}_1}^\top \hat{\mathbf{g}}_2),
\end{equation*}

\noindent where $\mathbf{R}_{\hat{\mathbf{g}}_1} \in \mathrm{SO}(3)$ is a rotation matrix that aligns $\hat{\mathbf{g}}_1$ with the vertical axis, ensuring that perturbations remain tangent to the manifold. The explicit expressions are provided in~\cite{teach-and-repeat}.
This parametrization enables a minimal and numerically stable representation of gravity direction perturbations while preserving the unit-norm constraint. Unit vectors $\hat{\mathbf{g}}_i$ can be locally parameterized using 2D perturbations in the tangent space at $\hat{\mathbf{g}}$. The mapping from a local perturbation $\boldsymbol{\delta} \in \mathbb{R}^2$ to a variation in the ambient space $\mathbb{R}^3$ is given by the Jacobian $\mathbf{J}_{\mathbb{S}^2} \in \mathbb{R}^{3 \times 2}$. This Jacobian is constructed from an orthonormal basis $\mathbf{B}$ of the tangent space, such that $\mathbf{J}_{\mathbb{S}^2} = \mathbf{B}$, with
\[
\hat{\mathbf{g}}^\top \mathbf{B} = \mathbf{0}, \quad \mathbf{B}^\top \mathbf{B} = \mathbf{I}_2.
\]

This construction ensures that perturbations remain on the tangent plane and respect the unit-norm constraint, in line with standard manifold-based estimation strategies~\cite{barfoot2014associating, barrau2017invariant}. 

\section{GNSS-inertial Initialization}
\label{sec:our-approach}

\begin{figure}[]
    \centering
    {\includegraphics[width=0.90\linewidth]{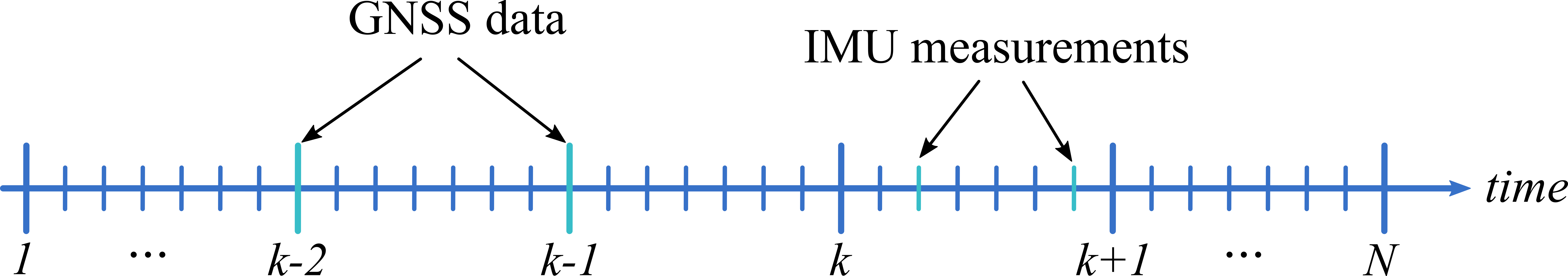}}
\caption{Illustration of the temporal notation for IMU and GNSS measurements.}
\label{linea-temporal}
\end{figure}

Our goal is estimating the state $\mathbf{x}$ of a sensing platform equipped with GNSS and inertial sensors. 
The information from both arrive at different rates, as shown in Fig. \ref{linea-temporal}.
The state consists of the device's rotation, position, linear velocity, IMU gyroscope bias, and gravity direction at different time steps.
We define the general state $\mathbf{x}$ as

\begin{equation}
        \mathbf{x} = \left[\mathbf{x}_1^\top,\mathbf{x}_2^\top, \cdots, \mathbf{x}_k^\top, \cdots  ,\mathbf{x}_N^\top, {\mathbf{b}^g}^\top, {\hat{\mathbf{g}}}^\top\right]^\top~,
\end{equation}
\noindent where
\begin{equation}
        \mathbf{x}_k = \left[\mathbf{R}_k^\top, 
 {\mathbf{p}}_k^\top,{\boldsymbol{\omega}}_k^\top,{\mathbf{v}}_k^\top\right]^\top~,
\end{equation}

\noindent and $\mathbf{R}_k \in SO(3)$ denotes the rotation matrix, $\mathbf{p}_k \in \mathbb{R}^3$ the position, $\boldsymbol{\omega}_k \in \mathbb{R}^3$ the angular velocity, $\mathbf{v}_k \in \mathbb{R}^3$ the linear velocity,  and $\mathbf{b}^g \in \mathbb{R}^3$ the gyroscope bias.
We assume a known gravity magnitude \(\|\mathbf{g}\| = 9.81~\mathrm{m/s^2}\) and only optimize the gravity direction (See Sec. \ref{subsec:gravity-representation}). The optimization is initialized with the identity quaternion, and the short initial window jointly refines attitude, gravity, and IMU biases; IMU rotational increments together with GNSS inter-epoch baselines provide sufficient yaw information for convergence without requiring an external heading prior.
It is worth noting that the proposed method does not require high-rate GNSS measurements. 
The key requirement is the availability of inter-epoch baselines that are sufficiently informative, which depends on platform motion rather than on GNSS sampling frequency. 

\subsection{Cost function formulation}
\label{subsec:cost-function-formulation}

We formulate a nonlinear optimization problem over the state variables $\mathbf{x}$, incorporating constraints from inertial preintegration and GNSS measurements. The objective is to find the optimal estimate $\hat{\mathbf{x}}$ that minimizes a joint cost function $J(\mathbf{x})$:
\begin{equation}
\label{eq:optimizacion_problem}
\hat{\mathbf{x}} = \argmin_{\mathbf{x}} J(\mathbf{x}).
\end{equation}
This cost function aggregates several residual terms, each encoding information from a different sensor modality or physical constraint. 
We begin with the angular velocity residual $\mathbf{r}_{\omega_k} \in \mathbb{R}^3$, which captures the discrepancy between the measured and predicted gyroscope readings. Assuming additive Gaussian noise $\boldsymbol{\eta}^g \sim \mathcal{N}(\textbf{0}, \boldsymbol{\Sigma}_{\omega})$, we define:
\begin{equation}
\label{eq:angular-velocity-term}
\mathbf{r}_{\omega_k} = \, ^{B}\boldsymbol{\omega}_{k} - \left(^{B}\Tilde{\boldsymbol{\omega}}_{k} - \mathbf{b}_k^g\right)
\end{equation}

Next, we introduce the IMU preintegration residuals, following the formulation of~\cite{forster-manifold-2017}. These include the rotation, velocity, and position residuals $\mathbf{r}_{\Delta \mathbf{R}}, \mathbf{r}_{\Delta \mathbf{v}}, \mathbf{r}_{\Delta \mathbf{p}} \in \mathbb{R}^3$:
\begin{align}
\mathbf{r}_{\Delta \mathbf{R}} &= \mathrm{Log}\left( \Delta \Tilde{\mathbf{R}}_{ij}^\top  \mathbf{R}_i^\top \mathbf{R}_j \right) \\
\mathbf{r}_{\Delta \mathbf{v}} &= \mathbf{R}_i^\top(\mathbf{v}_j - \mathbf{v}_i - \mathbf{g}\Delta t_{ij}) - \Delta \Tilde{\mathbf{v}}_{ij} \\
\mathbf{r}_{\Delta \mathbf{p}} &= \mathbf{R}_i^\top(\mathbf{p}_j - \mathbf{p}_i - \mathbf{v}_i \Delta t_{ij} - \tfrac{1}{2} \mathbf{g} \Delta t_{ij}^2) - \Delta \Tilde{\mathbf{p}}_{ij}
\end{align}

Each of these residuals is weighted by its corresponding covariance matrix, denoted ${\boldsymbol\Sigma}_{\Delta \boldsymbol\phi_{}}, \boldsymbol{\Sigma}_{\Delta \mathbf{v}}, \boldsymbol{\Sigma}_{\Delta \mathbf{p}} \in \mathbb{R}^{3 \times 3}$, as described in Section~\ref{subsec:IMU-model}.

To incorporate GNSS constraints, we consider the following measurement model:
\begin{equation*}
\hat{\mathbf{p}}^{\text{g}}_k = \mathbf{p}_k + \mathbf{n}_k, \quad \mathbf{n}_k \sim \mathcal{N}(\mathbf{0}, \boldsymbol{\Sigma}_{\text{p}})
\end{equation*}


Our formulation will use either global GNSS residuals when fully initialized, and inter-epoch GNSS baseline residuals initially. Accounting for the physical offset between the IMU and the GNSS antenna, the global residuals $\mathbf{r}_{\text{p}}\in \mathbb{R}^{3}$ are defined as:
\begin{equation}
\mathbf{r}_{\text{p}} = \mathbf{p}_k + \mathbf{R}_k \,\mathbf{l}_{B}^{\text{GNSS}} - \hat{\mathbf{p}}_k^{\text{g}},
\end{equation}
where $\mathbf{l}_{B}^{\text{GNSS}}$ is the fixed lever-arm vector from the IMU to the GNSS antenna expressed in the body frame.
Inter-epoch GNSS baseline vector refers to the 3D displacement of the antenna phase center between two time instants. 
Inter-epoch GNSS baseline vector residuals $\mathbf{r}_{\text{d}}\in \mathbb{R}^{3}$ enforce local trajectory consistency by inter-frame motion constraints between time instants $j$ and $k$.
To account for the physical offset between the IMU and the GNSS antenna, the lever arm is included in both poses, as follows
\begin{equation}
\mathbf{r}_{\text{d}} =
\left[ \mathbf{p}_j + \mathbf{R}_j \,\mathbf{l}_{B}^{\text{GNSS}} - \mathbf{p}_k - \mathbf{R}_k \,\mathbf{l}_{B}^{\text{GNSS}} \right]
- \left( \hat{\mathbf{p}}_j^{\text{g}} - \hat{\mathbf{p}}_k^{\text{g}} \right),
\end{equation}

Following~\cite{groves2008}, throughout this work, we assume that the GNSS–IMU lever arm $\mathbf{l}_{B}^{\text{GNSS}}\!\in\!\mathbb{R}^3$ is pre-calibrated. 

The global GNSS error is propagated to the inter-epoch baseline residuals as  $\boldsymbol{\Sigma}_{\text{d}} = 2 \boldsymbol{\Sigma}_{\text{p}}$.

Additionally, we incorporate a gravity residual $\mathbf{r}_{\mathbf{g}} \in \mathbb{R}^3$, based on the discrepancy between the estimated acceleration and the known gravity direction. Following the model introduced in Section~\ref{subsec:gravity-representation}, we define:
\begin{equation}
\label{eq:gravity-residual}
\mathbf{r}_{\mathbf{g}} = \left ( 
\frac{\mathbf{v}_j - \mathbf{v}_i}{\Delta t_{ij}} -
\frac{\mathbf{R}_i \Delta {\mathbf{v}}_{ij}}{\Delta t_{ij}} 
\right ) - \left (\|\mathbf{g}\| \hat{\mathbf{g}}\right ),
\end{equation}
where $\hat{\mathbf{g}} \in {S}^2$ denotes the estimated gravity direction and $\|\mathbf{g}\| = 9.81~\mathrm{m/s}^2$ is the known gravity magnitude. 
Assuming the dominant uncertainty arises from inertial preintegration, the associated covariance is approximated as $\boldsymbol{\Sigma}_{\mathbf{g}} = {\boldsymbol{\Sigma}_{\Delta \mathbf{v}}}/{\Delta t_{ij}^2}$.

Finally, the overall cost function is constructed as a sum of squared Mahalanobis norms. During initialization, the cost function is

\begin{dmath*}
\label{eq:cost-function-i}
J_i(\mathcal{X}) = \sum_{i=1}^{N} \left(
\left\| \mathbf{r}_{\Delta \mathbf{p},i} \right\|^2_{\boldsymbol{\Sigma}_{\Delta \mathbf{p}}} +
\left\| \mathbf{r}_{\Delta \mathbf{v},i} \right\|^2_{\boldsymbol{\Sigma}_{\Delta \mathbf{v}}} +
\left\| \mathbf{r}_{\omega,i} \right\|^2_{\boldsymbol{\Sigma}_{\omega}} +
\left\| \mathbf{r}_{\mathbf{g},i} \right\|^2_{\boldsymbol{\Sigma}_{\mathbf{g}}} +
\left\| \mathbf{r}_{\Delta \mathbf{R},i} \right\|^2_{{\boldsymbol\Sigma}_{\Delta \boldsymbol\phi_{}}} \right) +
\sum_{j,k \in S}\left\| \mathbf{r}_{\text{d},j, k} \right\|^2_{\boldsymbol{\Sigma}_{\text{d}}}
\end{dmath*}

\noindent where $N$ denotes the number of GNSS measurements included in the optimization, and $S$ the set of GNSS pairs considered for the distance constraints. After the initialization criterion is met, the cost function is switched to the following one, that replaces inter-epoch baseline GNSS residuals by the global GNSS ones

\begin{dmath*}
\label{eq:cost-function}
J(\mathcal{X}) = \sum_{i=1}^{N} \left(
\left\| \mathbf{r}_{\Delta \mathbf{p},i} \right\|^2_{\boldsymbol{\Sigma}_{\Delta \mathbf{p}}} +
\left\| \mathbf{r}_{\Delta \mathbf{v},i} \right\|^2_{\boldsymbol{\Sigma}_{\Delta \mathbf{v}}} +
\left\| \mathbf{r}_{\omega,i} \right\|^2_{\boldsymbol{\Sigma}_{\omega}} +
\left\| \mathbf{r}_{\mathbf{g},i} \right\|^2_{\boldsymbol{\Sigma}_{\mathbf{g}}} +
\left\| \mathbf{r}_{\Delta \mathbf{R},i} \right\|^2_{{\boldsymbol\Sigma}_{\Delta \boldsymbol\phi_{}}} +
\left\| \mathbf{r}_{\text{p},i} \right\|^2_{\boldsymbol{\Sigma}_{\text{p}}}
\right)
\end{dmath*}

\begin{figure*}[]
\footnotesize
\centering
\begin{equation*}
\label{eq:obs-matrix-full}
\mathcal{O} =
\left[
\begin{array}{cccccccc}
-\mathbf{J}_r^{-1}(\boldsymbol{\phi})
 \mathbf{R}_j ^\top \mathbf{R}_i & \mathbf{0} & \mathbf{0} &
\mathbf{J}_r^{-1}(\boldsymbol{\phi})
 & \mathbf{0} & \mathbf{0} &
\mathbf{0} & \mathbf{0}
\\[1.5ex]
\left( \mathbf{R}_i^\top (\mathbf{v}_j - \mathbf{v}_i - \|\mathbf{g}\| \hat{\mathbf{g}} \Delta t_{ij}) \right)^\wedge &
-\mathbf{R}_i^\top & \mathbf{0} &
\mathbf{0} & \mathbf{R}_i^\top & \mathbf{0} &
\mathbf{0} & -\|\mathbf{g}\| \mathbf{R}_i^\top \Delta t_{ij}
\\[1.5ex]
\left( \mathbf{R}_i^\top (\mathbf{p}_j - \mathbf{p}_i - \mathbf{v}_i \Delta t_{ij} - \tfrac{1}{2} \|\mathbf{g}\| \hat{\mathbf{g}} \Delta t_{ij}^2) \right)^\wedge &
-\mathbf{R}_i^\top \Delta t_{ij} & -\mathbf{R}_i^\top &
\mathbf{0} & \mathbf{0} & \mathbf{R}_i^\top &
\mathbf{0} & -\tfrac{1}{2} \|\mathbf{g}\| \mathbf{R}_i^\top \Delta t_{ij}^2
\\[1.5ex]
+\;\mathbf{R}_i\,[\mathbf{l}_B^{\text{GNSS}}]^\wedge  & \mathbf{0} & -\mathbf{I}_3 &
-\;\mathbf{R}_j\,[\mathbf{l}_B^{\text{GNSS}}]^\wedge & \mathbf{0} & \mathbf{I}_3 &
\mathbf{0} & \mathbf{0}
\\[1.5ex]
-\;\mathbf{R}_i\,[\mathbf{l}_B^{\text{GNSS}}]^\wedge & \mathbf{0} & \mathbf{I}_3 &
\mathbf{0} & \mathbf{0} & \mathbf{0} &
\mathbf{0} & \mathbf{0}
\\[1.5ex]
\mathbf{0} & \mathbf{0} & \mathbf{0} &
-\;\mathbf{R}_j\,[\mathbf{l}_B^{\text{GNSS}}]^\wedge & \mathbf{0} & \mathbf{I}_3 &
\mathbf{0} & \mathbf{0}
\\[1.5ex]
\mathbf{0} & \mathbf{0} & \mathbf{0} &
\mathbf{0} & \mathbf{0} & \mathbf{0} &
\mathbf{I}_3 & \mathbf{0}
\\[1.5ex]
\mathbf{0} & -\tfrac{1}{\Delta t_{ij}} \mathbf{I}_3 & \mathbf{0} &
\mathbf{0} & \tfrac{1}{\Delta t_{ij}} \mathbf{I}_3 & \mathbf{0} &
\mathbf{0} & -\|\mathbf{g}\| \mathbf{J}_{\mathbb{S}^2}
\end{array}
\right]
\end{equation*}
\caption{Structure of the full observability matrix $\mathcal{O}$, obtained by stacking the Jacobians of all residuals with respect to the state vector $\mathbf{x}$. The Jacobian $\mathbf{J}_r^{-1}(\boldsymbol{\phi})$ denotes the inverse of the right Jacobian of $\mathrm{SO}(3)$, evaluated at the rotational error $\boldsymbol{\phi} = \mathrm{Log} ( \Delta \Tilde{\mathbf{R}}_{ij}^\top \mathbf{R}_i^\top \mathbf{R}_j )$. The last row accounts for the gravitational constraint, where the gravity direction $\hat{\mathbf{g}} \in \mathbb{S}^2$ is parametrized on the unit sphere and its Jacobian is denoted by $\mathbf{J}_{\mathbb{S}^2}$. Accounting for the GNSS–inertial lever arm makes GNSS residuals orientation-sensitive. Absolute terms inform attitude at their node and relative baselines at both endpoints, boosting yaw/bias observability under rotation while reducing to position-only if the arm is negligible.
}
\label{fig:obs-matrix-explicit}
\end{figure*}
\subsection{Observability Analysis}
\begin{figure}[]
\centering
\includegraphics[width=0.79\columnwidth]{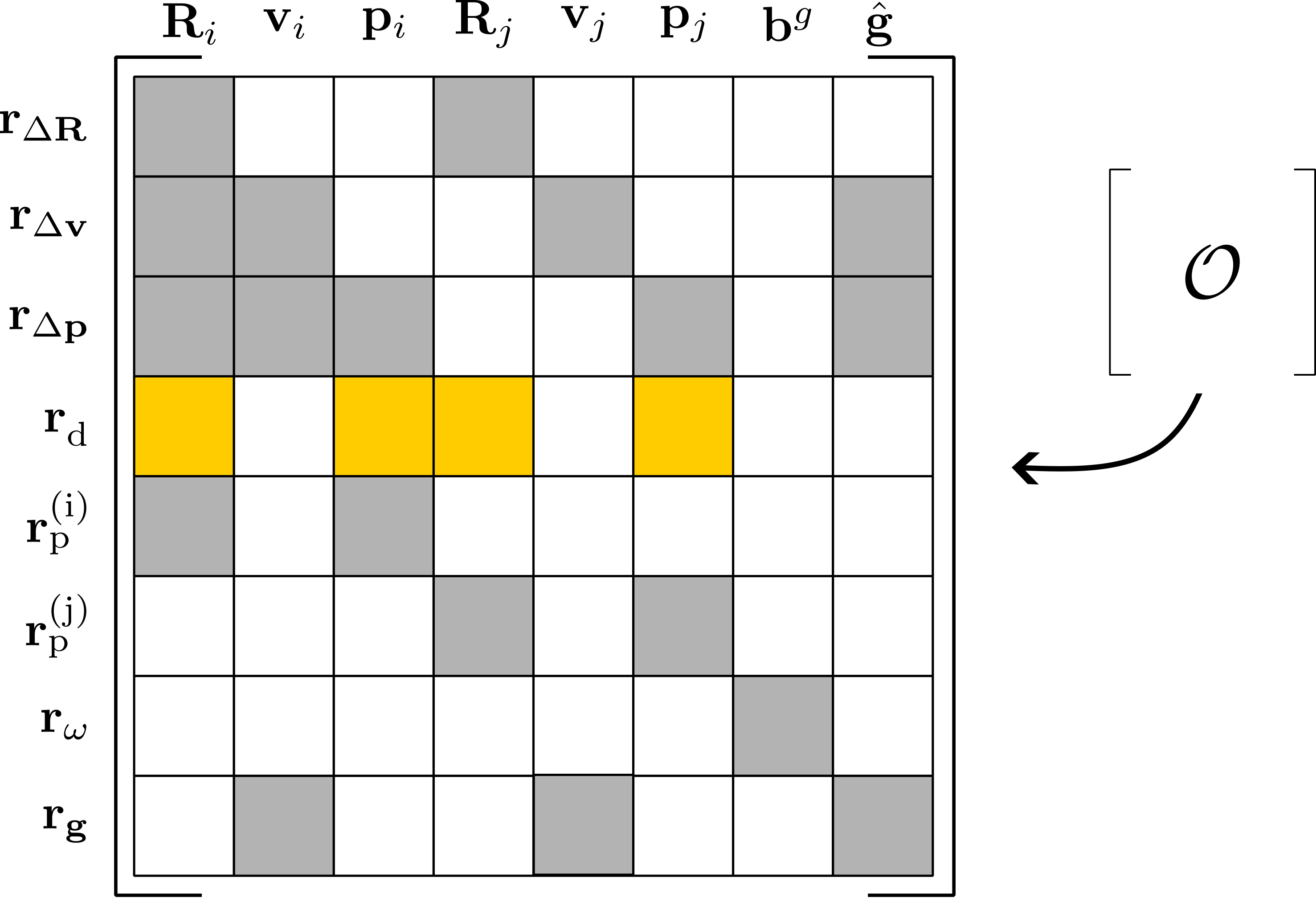}
\caption{
Structure of the observability matrix $\mathcal{O}$.
Each row corresponds to a residual and each column to a component of the state vector.
White cells indicate matrices with zero values, meaning no sensitivity to that state.
Gray cells represent general nonzero Jacobians that contribute to observability.
Orange cells denote relative constraints that do not improve global observability.
}
\label{fig:graphical-observability_matrix}
\end{figure}

Observability in nonlinear systems is traditionally analyzed through successive Lie derivatives of the output function $h(\mathbf{x})$ with respect to the system dynamics $\dot{\mathbf{x}} = f(\mathbf{x})$. This process leads to the construction of a nonlinear observability matrix by stacking gradients such as $\nabla h(\mathbf{x})$, $\nabla \mathcal{L}_f h(\mathbf{x})$, $\nabla \mathcal{L}_f^2 h(\mathbf{x})$, and higher-order terms~\cite{hermann1977nonlinear}. 
In this work, we adopt a discrete-time perspective in which observability is assessed by linearizing the measurement residuals over multiple time steps. Although Lie derivatives are not computed explicitly, the Jacobians of these residuals capture the system's sensitivity to its internal state in a manner analogous to the continuous-time formulation. This strategy has been studied and justified in~\cite{hesch2014consistency,huang2009observability}, where it is shown that the rank of the stacked Jacobian matrix determines the locally observable directions and directly affects estimator consistency.
Therefore, we apply this methodology to analyze the observability of our system within a discrete-time factor graph formulation. Specifically, we consider a temporal window spanning two consecutive time steps, denoted as $\{t_i, t_j\}$, and define the corresponding state vector as
\begin{equation}
\label{eq:obs-vector-state}
\mathbf{x} = \left[ \mathbf{R}_i, \mathbf{v}_i, \mathbf{p}_i,\; \mathbf{R}_j, \mathbf{v}_j, \mathbf{p}_j,\; \mathbf{b}^g, \hat{\mathbf{g}} \right] ~\in \mathbb{R}^{23}.
\end{equation}

To analyze the system’s observability, we construct the observability matrix $\mathcal{O}$ by stacking the Jacobians of all residuals in the cost function with respect to the state vector  (\ref{eq:obs-vector-state}), as follows
\begin{align}
\mathcal{O} &= \notag \\[-1.5ex]
&\scalebox{0.95}{$
\begin{bmatrix}
\frac{\partial \mathbf{r}_{\Delta \mathbf{R}}}{\partial \mathbf{x}}^\top,
\frac{\partial \mathbf{r}_{\Delta \mathbf{v}}}{\partial \mathbf{x}}^\top,
\frac{\partial \mathbf{r}_{\Delta \mathbf{p}}}{\partial \mathbf{x}}^\top,
\frac{\partial \mathbf{r}_{\text{d}}}{\partial \mathbf{x}}^\top,
\frac{\partial \mathbf{r}_{\text{p}_i}}{\partial \mathbf{x}}^\top,
\frac{\partial \mathbf{r}_{\text{p}_j}}{\partial \mathbf{x}}^\top,
\frac{\partial \mathbf{r}_{\omega}}{\partial \mathbf{x}}^\top,
\frac{\partial \mathbf{r}_{\mathbf{g}}}{\partial \mathbf{x}}^\top
\end{bmatrix}^\top
$}
\label{eq:observability_matrix}
\end{align}

Based on the residual definitions introduced in Section~\ref{subsec:cost-function-formulation}, the explicit expression of the observability matrix~$\mathcal{O}$ is shown in Fig.~\ref{fig:obs-matrix-explicit}, while Fig.~\ref{fig:graphical-observability_matrix} illustrates its structure graphically: each row corresponds to a residual and each column to a state variable. Nonzero entries are marked in color, with gray indicating general contributions and orange highlighting relative-only constraints.

A symbolic evaluation shows that $\mathcal{O}$ has a maximum rank of 23. This is lower than the number of columns because the GNSS relative constraint is linearly dependent on the pair of absolute constraints, contributing six independent equations instead of nine. To resolve this underconstrained setup during initialization, we assume the first pose is known. Fixing position, velocity, and orientation at $t_0$ reduces the number of unknowns from 23 to 14. In this reduced system, $\mathcal{O}$ achieves rank 11, with the unobservable directions mainly linked to the second pose, gyroscope bias, and gravity orientation.

The actual rank of $\mathcal{O}$ depends on platform motion. In constant-velocity, straight-line trajectories, the accelerometer measures only gravity, making accelerometer bias and gravity indistinguishable. Likewise, without rotational excitation, yaw and gyroscope bias cannot be reliably separated. In both cases, the associated Jacobian columns become linearly dependent, reducing the rank of $\mathcal{O}$.
This can be seen directly from our residual formulation. The velocity and position preintegration residuals ($\mathbf{r}_{\Delta v}, \mathbf{r}_{\Delta p}$) contain both accelerometer bias and gravity terms ($-\mathbf{R}_i^\top$ and $-\|\mathbf{g}\|\mathbf{R}_i^\top\Delta t_{ij}$). With nearly constant attitude and no non-gravitational accelerations, these Jacobians collapse, causing the two states to be inseparable. Similarly, the angular velocity residual $\mathbf{r}_{\omega_k}$ and the rotational residual $\mathbf{r}_{\Delta R}$ constrain yaw and gyroscope bias only if the platform undergoes rotations; otherwise, their Jacobians degenerate and yaw couples with $\mathbf{b}_g$. As a result, certain motion patterns manifest as rank deficiencies in Fig.~\ref{fig:obs-matrix-explicit}. Informative maneuvers combining accelerations and rotations are therefore essential to restore full observability~\cite{groves2008}.

Our approach does not rely on external priors for orientation or gravity. Instead, it leverages the coupling between inertial preintegration and GNSS measurements to progressively constrain the state. The simplifying assumptions at initialization are temporary: as the estimation window $\{t_i,t_j\}$ advances and new measurements overlap, additional constraints accumulate and observability improves. This progressive recovery, also reported in~\cite{eckenhoff2018preintegration}, motivates our criterion that delays the activation of global GNSS residuals until the system has been sufficiently excited, thereby avoiding premature anchoring to noisy absolute positions and improving initialization robustness.

\subsection{Triggering Criterion for including Global Residuals}
\label{sec:activation-criterion}

\begin{figure}[]
    \centering
    {\includegraphics[width=0.91\linewidth]{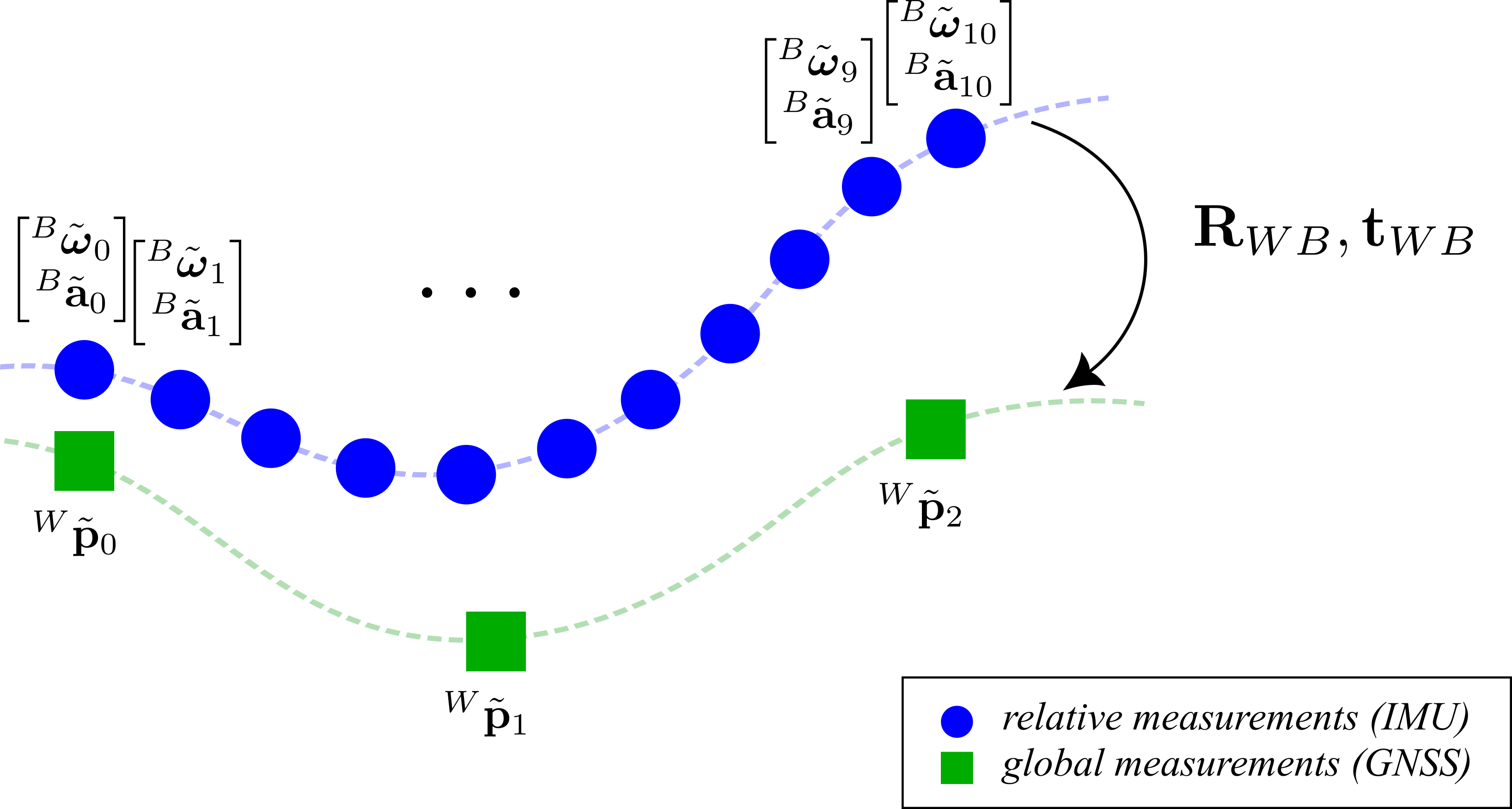}}
    \caption{Initialization with GNSS measurements are shown as green squares, while IMU data are represented as blue circles. Alignment ensures that relative motion inferred from the IMU is consistent with the global trajectory.}
    \label{fig:imu_gps_alignment}
\end{figure}
While the IMU provides relative motion information, GNSS delivers absolute position measurements.
In order to effectively integrate these two sources of information, a rigid transformation denoted as $\T_{WB}{} \in SE\parenthesis{3}$ is employed. 
Fig.~\ref{fig:imu_gps_alignment} illustrates the aforementioned concept, depicting GNSS measurements as green squares and IMU data points as blue circles. While GNSS provides globally-referenced position information, it typically suffers from higher noise levels. In contrast, IMU measurements are available with lower short-term noise, but they are subject to significant drift over time due to biases and integration errors. 
To ensure a robust estimation of $\T_{WB}{}$, we postpone its optimization until a sufficient number of GNSS measurements have been accumulated. This delay increases the system’s observability and reduces sensitivity to early-stage noise, as the integration of additional data improves excitation and allows for statistical averaging.

In order to determine the triggering time instant $k^*$, we evaluate the observability of the extrinsic transformation $\T_{WB}{}$ by analyzing the singular values of an approximate Hessian matrix built from GNSS-only residuals. These residuals include one relative term (i.e., displacement between consecutive GNSS positions) and two absolute anchor terms (i.e., global positions expressed in the world frame).
The Hessian matrix is obtained by linearizing the aforementioned residuals with respect to the six parameters of $\T_{WB}{}$, yielding the following approximation:
\begin{equation}
    \mathbf{H} \approx \sum_{k=1}^{N} \mathbf{J}_k^\top \boldsymbol{\Omega}_k \mathbf{J}_k,
\end{equation}
\noindent
where $\mathbf{J}_k \in \mathbb{R}^{9 \times 6}$ is the Jacobian of the residuals at time step $k$ with respect to $\T_{WB}{}$, and $\boldsymbol{\Omega}_k = \boldsymbol{\Sigma}_k^{-1} \in \mathbb{S}^{9}_+$ is the inverse covariance matrix of the stacked GNSS measurement noise.
To assess the structure and conditioning of the optimization problem, we compute the singular value decomposition (SVD) of the resulting Hessian $\mathbf{H} \in \mathbb{S}^{6}_+$. From this decomposition, we obtain a set of singular values and define the singular value ratio as $\rho_k = \sigma_{\max}/\sigma_{\min}$, where $\sigma_{\min}$ and $\sigma_{\max}$ denote the smallest non-zero and the largest singular values of $\mathbf{H}$, respectively.
Rather than incorporating global residuals prematurely, we defer their inclusion until the estimation problem becomes sufficiently constrained. To decide when to activate global alignment, we monitor the evolution of the relative change in the singular value ratio:

\begin{equation}
    \Delta \rho_k = \left| \frac{\rho_k - \rho_{k-1}}{\rho_{k-1}} \right|.
\end{equation}

A large value of $\Delta \rho_k$ indicates that the observability of the system is still evolving, whereas consistently low values suggest that the structure of the Hessian has stabilized and the states are observable. Based on this, we introduce global GNSS residuals and begin estimating the transformation $\T_{WB}{}$ only when the system becomes sufficiently constrained. Specifically, we activate these terms once the following condition is met:
\begin{equation}
\Delta \rho_k < \Delta \rho_{\mathrm{th}}~,
\end{equation}
where $\Delta \rho_{\mathrm{th}}$ is a fixed threshold. The index $k^*$ denotes the time at which this criterion is satisfied. This activation rule complements our observability analysis by providing a concrete and data-driven signal for when global residuals can be safely introduced. In contrast to prior methods that rely on heuristic indicators or manually tuned thresholds~\cite{xiao2020observability, huang2009observability}, our approach monitors the evolution of the Hessian’s conditioning to make this decision autonomously.

\section{Experiments and Results}
\label{sec:experiments}

\subsection{Experimental setup}

All experiments were run on an Intel Core i7-11700K CPU (3.6 GHz, 64 GB RAM) 
using the open-source \texttt{PyPose} library~\cite{wang2023pypose} for Lie group operations and optimization.
We assess tracking performance using the Root Mean Square of the Absolute Trajectory Error (ATE RMSE~$\downarrow$) 
Besides trajectory accuracy, we report RMSE of gyro bias estimates and heading (yaw) error after initialization, following~\cite{Shin2005EstimationTechniques}.
The EuRoC and GVINS datasets were selected as standard benchmarks in GNSS–inertial fusion~\cite{cioffi2020tightly}.
We additionally evaluate on the MARS LVIG dataset, which contains large-scale outdoor trajectories. Sequences collected including manual flights at a low-altitude ($20~\mathrm{m}$) were used in our experiments. Unlike EuRoC and GVINS, it features long GNSS outages and stronger dynamics, providing a more challenging benchmark for testing the robustness of our initialization.

\subsection{Synthetic experiments}

We conduct two complementary analyses on synthetic GNSS data, designed to isolate and better understand the influence of different modeling choices within our optimization framework. 

First, we evaluate two alternative formulations of GNSS residuals within the same nonlinear optimizer and under a fixed-size sliding window. In the \textit{absolute} variant (A1), we introduce absolute position constraints at each epoch, defined as \(r_k^{\mathrm{abs}} = \mathbf{p}^{\mathrm{GNSS}}_k - \mathbf{p}_k(\mathbf{x})\). In the \textit{inter-epoch} variant (A2), we instead enforce consistency between consecutive GNSS baselines, with residuals \(r_k^{\mathrm{base}} = (\mathbf{p}^{\mathrm{GNSS}}_k - \mathbf{p}^{\mathrm{GNSS}}_{k-1}) - (\mathbf{p}_k(\mathbf{x}) - \mathbf{p}_{k-1}(\mathbf{x}))\). Both use identical weights, robust loss, window length, and stopping criteria, ensuring a fair comparison. The evaluation is carried out on EuRoC by varying the GNSS noise level from $\sigma_p = 0.2$ to $2.0\,\mathrm{m}$, repeating each configuration ten times for statistical significance. 
\begin{table}[t]
\centering
\setlength{\tabcolsep}{5pt}
\resizebox{0.385\textwidth}{!}{%
\rowcolors{2}{white}{gray!15}
\begin{tabular}{l | ccc}
\toprule
$\sigma_{p}$ [$\mathrm{m}$] & A1 Pos. [$\downarrow \mathrm{m}$]  & A2 Pos. [$\downarrow \mathrm{m}$] & $\Delta\%$ (A2$-$A1) \\
\midrule
0.2 & 0.05 $\pm$ 0.01 & 0.05 $\pm$ 0.01 & 0.0 \\
1.0 & 0.53 $\pm$ 0.03 & \textbf{0.51 $\pm$ 0.04} & \textbf{$-3.8$} \\
2.0 & 1.64 $\pm$ 0.08 & \textbf{1.44 $\pm$ 0.08} & \textbf{$-12.2$} \\
\bottomrule
\end{tabular}}
\caption{EuRoC (mean$\,\pm\,$std over 10 runs). GNSS residuals: A1 = absolute position; A2 = inter-epoch baseline. $\Delta\%=\frac{\text{A2}-\text{A1}}{\text{A1}}\times 100$.}
\label{tab:ablation_a1_a2}
\end{table}
\begin{figure}[]
    \centering
    \includegraphics[width=\columnwidth]{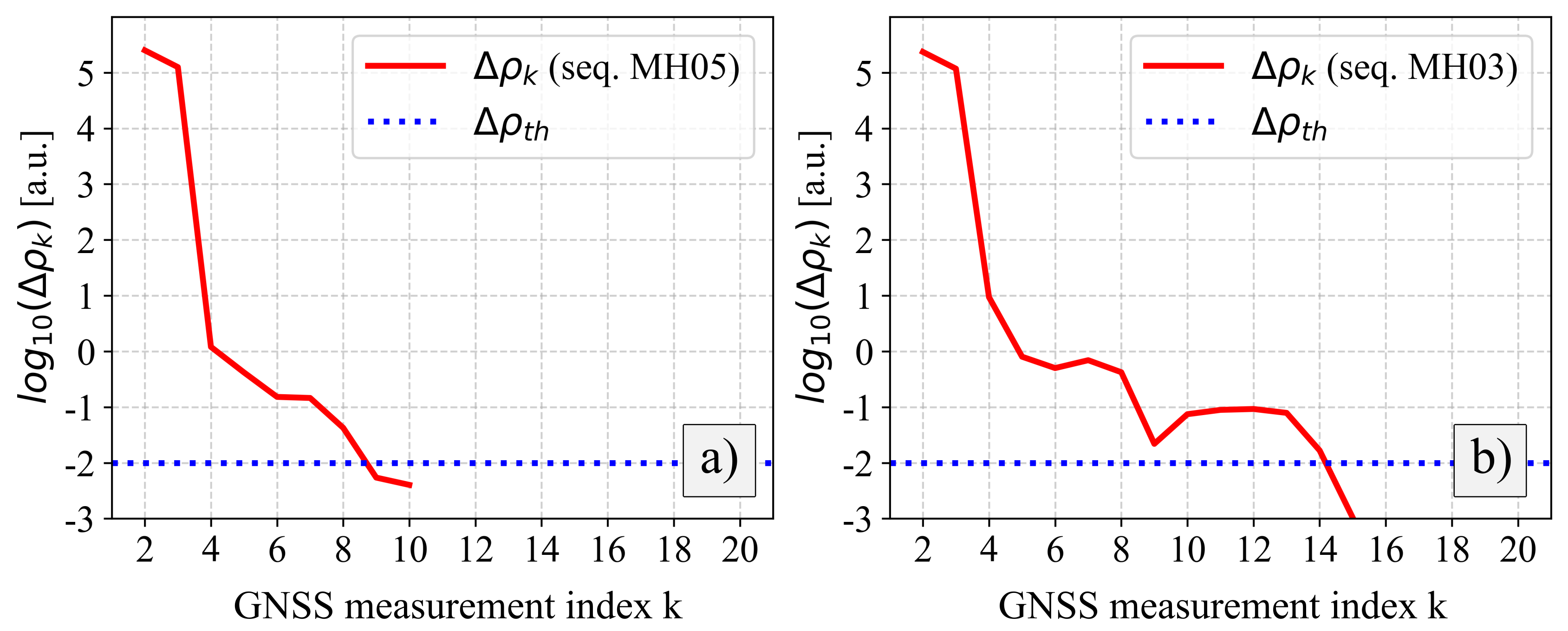}
    \caption{
    Example of our activation criterion for two EuRoc sequences. a) On the MH05 sequence, the criterion results in $k^*=9$ GNSS measurements; b) on the MH03 sequence, it suggests $k^*=15$ GNSS measurements. The dashed blue line represents the observability threshold $ \Delta \rho_{{th}}$.
    }
    \label{fig:activation-observability}
\end{figure}
As shown in Table~\ref{tab:ablation_a1_a2}, the two strategies yield similar performance under low-noise conditions. However, as GNSS quality deteriorates, the inter-epoch formulation (A2) proves more robust, leading to lower trajectory errors while maintaining essentially unchanged heading and gyroscope-bias estimates. These results suggest that inter-epoch baseline residuals are better suited for GNSS-challenged regimes, a design choice further validated with real GNSS data in the following subsections.

Second, we assess the effect of delaying the activation of global residuals and optimizing the rigid transformation $\T_{WB}$ using: (i) EuRoC MAV~\cite{burri2016euroc}, where GNSS is synthetically generated by adding zero-mean Gaussian noise with $\sigma_{\text{p}} = 0.2~\text{m}$ to motion-capture ground truth following~\cite{cioffi2020tightly}; and (ii) GVINS~\cite{wang2020gvins}, which provides real outdoor IMU and high-rate RTK-GNSS. For GVINS, we extract IMU/GNSS data and align positions to the UTM frame.

A central question in this setup is when to activate global position residuals. Incorporating them too early, while the system is weakly constrained or underconstrained, can degrade accuracy. Delaying their incorporation until sufficient GNSS measurements are available allows the optimizer to use them more effectively. To address this, we proposed an observability-based triggering criterion that determines the appropriate instant to introduce global constraints. Fig.~\ref{fig:activation-observability} illustrates the evolution of the relative change $\Delta \rho_k$ for two EuRoC sequences. When the curve drops below the observability threshold $\Delta \rho_{\mathrm{th}} = 10^{-2}$, the system is considered sufficiently constrained to support the inclusion of global residuals. For instance, the criterion suggests $k^*=9$ GNSS measurements for \textit{MH05} and $k^*=15$ for \textit{MH03}.

\begin{figure}[t]
    \centering
    {\includegraphics[width=\linewidth]{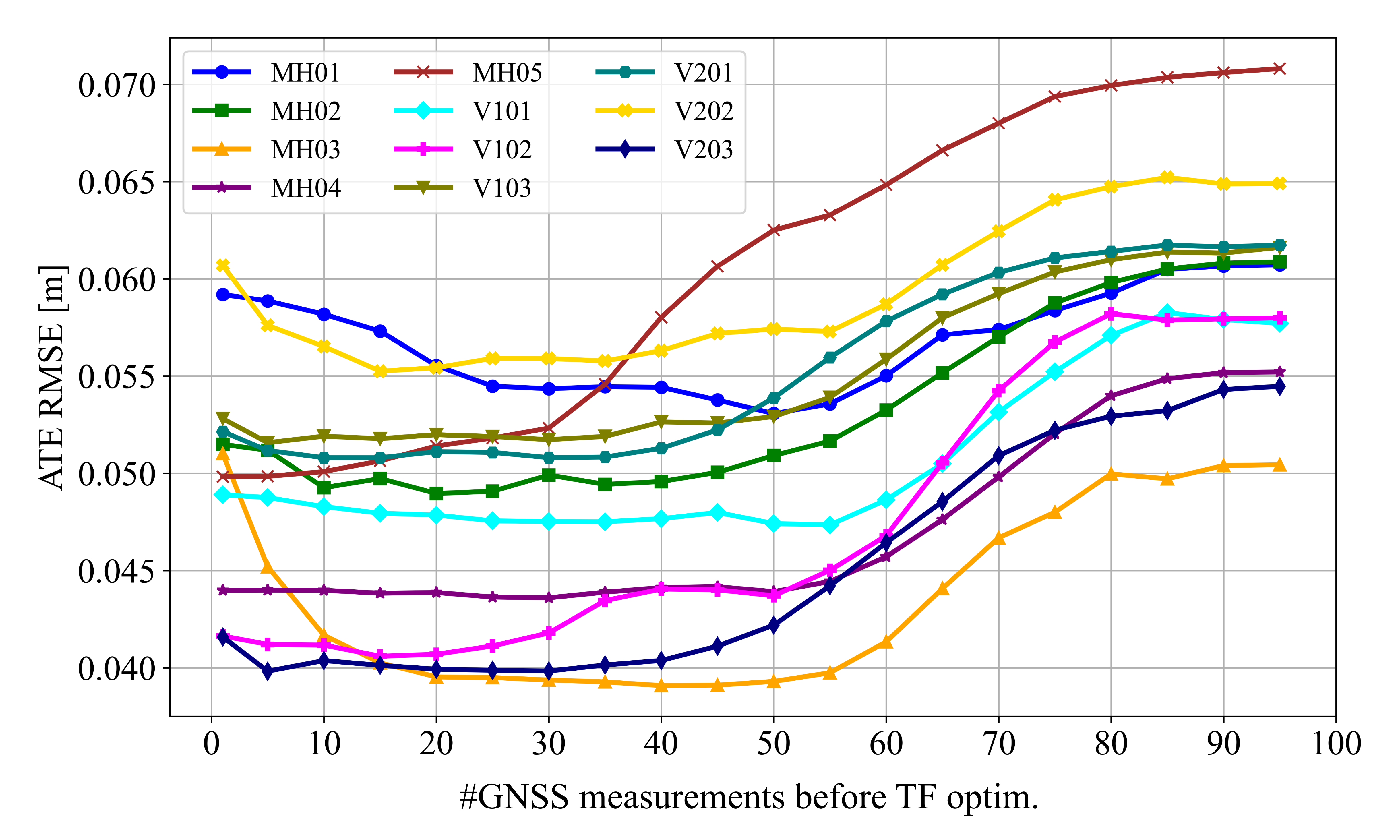}}
    \caption{RMSE values for different EuRoc sequences. x-axis represent the number of GNSS measurements accumulated before including global residuals in the optimization.}
    \label{fig:euroc-position-errors}
\end{figure}

Finally, Fig.~\ref{fig:euroc-position-errors} shows the impact of this strategy on position accuracy. The x-axis in the figure denotes the number of GNSS measurements accumulated before enabling global residuals in the optimization. The full set of 100 measurements corresponds to a trajectory of approximately $20~\text{m}$. Activating global residuals prematurely increases error, whereas deferring them according to the proposed criterion leads to significantly lower trajectory error. These results highlight the importance of delaying global constraints to mitigate the effects of early noise and uncertainty.

\begin{table}[b]
    \centering
    \setlength{\tabcolsep}{2pt}
    \resizebox{0.485\textwidth}{!}{
    \rowcolors[]{2}{white}{gray!15}
    \begin{tabular}{l l | ccc }
        \toprule
        {Dataset} & {Sequence} & {$k^*-$index} & {Pos. Error (full)} & {Pos. Error (from $k^*$)} \\
        \midrule
        EuRoC & MH01        & 11 & 0.058 & \textbf{0.057} \\
              & MH02        & 12 & 0.051 & \textbf{0.049} \\
              & MH03      & 15 & 0.051 & \textbf{0.040} \\
              & MH04   & 21 & 0.044 & \textbf{0.043} \\
              & MH05   & 9  & \textbf{0.049} & 0.050 \\
              & V101        & 20 & 0.048 & \textbf{0.046} \\
              & V102      & 10 & 0.041 & \textbf{0.040} \\
              & V103   & 30 & 0.053 & \textbf{0.052} \\
              & V201        & 21 & 0.051 & \textbf{0.050} \\
              & V202     & 10 & 0.061 & \textbf{0.057} \\
              & V203   & 12 & 0.042 & \textbf{0.040} \\
        \midrule
        GVINS & Sports field       & 19  & 0.52 &\textbf{0.50} \\
              & Indoor-outdoor     & 49 & 0.52 & \textbf{0.50} \\
              & Urban driving      & 10 & 0.54 & \textbf{0.53} \\
        \bottomrule
    \end{tabular}
    }
    \caption{
         $k^*-$index obtained by our observability criterion and position errors. 
        `Pos. Error (Full)' refers to ATE RMSE [$\downarrow{\mathrm{m}}$] when using GNSS global residuals from the outset, while 
        `Pos. Error (from $k^*$)' shows results when global residuals are introduced at  $k^*$. Note consistent improvements by delaying global residuals until $k^*$.
    }
    \label{tab:activation-index}
\end{table}
Table~\ref{tab:activation-index} extends this analysis by reporting the activation index $k^*$ and the resulting errors across EuRoC and GVINS. Our method consistently achieves lower position errors than the baselines, with the largest gains in sequences such as \textit{MH\_03} and \textit{Indoor-outdoor}. On average, trajectory error is reduced by about $4\%$, reaching up to $20\%$ in specific cases. These results validate the observability-based activation strategy, showing that postponing the inclusion of global residuals until the system is sufficiently constrained improves both accuracy and robustness across diverse datasets.

\color{black}

\begin{table}[]
\centering
\setlength{\tabcolsep}{6pt}
\resizebox{0.46\textwidth}{!}{%
\rowcolors{5}{gray!15}{white}
\begin{tabular}{l | ccc | ccc }
\toprule
\multicolumn{1}{c|}{ } &
\multicolumn{3}{c|}{Heading Err. [$\downarrow \mathrm{deg}$]} &
\multicolumn{3}{c}{$\mathbf{b}^g$ Err. [$\downarrow \mathrm{rad/s}$]} \\
\cmidrule(lr){2-4}\cmidrule(lr){5-7}
Sequence & EKF & RA-JAPE & Ours & EKF & RA-JAPE & Ours \\
\midrule
airport1 & 1.64 & 1.61 & \textbf{1.59} & 0.048 & 0.044 & \textbf{0.040} \\
airport2 & 1.73 & 1.70 & \textbf{1.68} & 0.052 & 0.048 & \textbf{0.044} \\
airport3 & 1.79 & 1.78 & \textbf{1.75} & 0.052 & \textbf{0.044} & 0.045 \\
island1  & \textbf{1.30} & 1.31 & 1.32 & 0.040 & 0.040 & \textbf{0.038} \\
island2  & 1.41 & 1.40 & \textbf{1.39} & 0.040 & 0.040 & \textbf{0.036} \\
island3  & 1.56 & 1.52 & \textbf{1.50} & 0.048 & 0.044 & \textbf{0.040} \\
town1    & 1.57 & 1.54 & \textbf{1.52} & 0.044 & 0.044 & \textbf{0.040} \\
town2    & 1.66 & \textbf{1.60} & 1.61 & 0.048 & \textbf{0.044} & 0.048 \\
town3    & 1.74 & 1.73 & \textbf{1.70} & 0.048 & 0.046 & \textbf{0.045} \\
\bottomrule
\end{tabular}%
}
\caption{MARS-LVIG data evaluation. Per-sequence heading [$\mathrm{deg}$] and gyroscope-bias [$\mathrm{rad/s}$] errors for EKF, RA-JAPE~\cite{wu2014new}, and Ours. Best per row is \textbf{bolded}.}
\label{tab:mars-table}
\end{table}

\subsection{MARS-LVIG experiments}
We further validate our approach on the MARS-LVIG dataset~\cite{li2024mars}, which provides multi-sensor UAV sequences with data collected across diverse large-scale outdoor scenes. We follow the same experimental setup, but in this case we report errors in heading and gyroscope bias, comparing our method against state-of-the-art techniques: EKF and RA-JAPE~\cite{wu2014new}. This evaluation allows us to assess not only the impact on position accuracy, but also the ability of the proposed activation strategy to improve attitude estimation and bias calibration under realistic field conditions.
Across the nine UAV sequences in Table~\ref{tab:mars-table}, the largest benefits appear on gyroscope bias: improvements reach up to +16.7\% versus EKF on \textit{airport1} and \textit{island3}, and +10.0\% versus RA-JAPE on \textit{island2}. Averaged across sequences, bias error decreases by +11\% versus EKF and +5\% versus RA-JAPE, with best results in 7/9 sequences. For heading, the best-case gains are +4\% versus EKF on \textit{island3} and +2\% versus RA-JAPE on \textit{town3}; averaged across sequences the gains are +3\% and +1\%, respectively, and we obtain the lowest heading in 6/9 sequences. 
A few reversals remain (e.g., \textit{island1}/\textit{island2} for heading, \textit{town2} and \textit{airport3} for bias), which we attribute to more challenging GNSS geometry or intermittent availability. While small in magnitude, such reductions are meaningful: sub-degree yaw errors at
initialization can accumulate into drift~\cite{Syed2008CivilianNavigation}. The improved
heading and bias estimates therefore yield a more reliable initial state for GNSS–inertial fusion. 

Overall, the trend indicates that inter-epoch baseline GNSS residuals with activation substantially improve bias estimation while maintaining competitive heading accuracy under realistic, large-scale flights.

\section{Conclusions}
\noindent 
In this paper, we introduced a GNSS-inertial initialization method formulated through inter-epoch baseline residuals. We also proposed a principled activation criterion that defers the inclusion of global residuals until the problem is sufficiently well-conditioned, as determined by an observability analysis. Experiments on  EuRoC, GVINS and MARS-LVIG datasets confirm that this strategy yields accurate and robust trajectory estimates under a variety of operating conditions. 
Furthermore, our approach generalizes effectively to realistic GNSS deployments, providing consistent improvements not only in position accuracy but also in heading and gyroscope-bias estimation. Taken together, these results validate the proposed method as a reliable initialization strategy across both controlled benchmarks and challenging real-world scenarios.

\color{black}




\end{document}